\documentclass[conference]{IEEEtran}
\IEEEoverridecommandlockouts
\usepackage{cite}
\usepackage{amsmath,amssymb,amsfonts}
\usepackage{algorithmic}

\usepackage{graphicx}
\usepackage[linesnumbered,ruled,vlined]{algorithm2e}

\def\BibTeX{{\rm B\kern-.05em{\sc i\kern-.025em b}\kern-.08em
    T\kern-.1667em\lower.7ex\hbox{E}\kern-.125emX}}

\usepackage[colorlinks=true,linkcolor=blue]{hyperref}%
\usepackage[ruled,linesnumbered]{algorithm2e}
\usepackage{soul,color}
\usepackage[utf8]{inputenc}
\usepackage[english]{babel}
\usepackage{textcomp}
\usepackage{xcolor}
\def\BibTeX{{\rm B\kern-.05em{\sc i\kern-.025em b}\kern-.08em
    T\kern-.1667em\lower.7ex\hbox{E}\kern-.125emX}}

\usepackage{tabularx}
\usepackage{booktabs}
\usepackage{lettrine}
\usepackage{multirow}

\usepackage{amssymb}

\usepackage{tabularx,colortbl}
\usepackage{mathtools}
\usepackage{eqnarray,amsmath}
\usepackage{amsmath}
\usepackage{amsmath,amssymb,amsfonts}
\usepackage{graphicx}
\usepackage{comment}
\usepackage{amssymb}
\usepackage{soul}
\usepackage[export]{adjustbox}

\usepackage[ruled,linesnumbered]{algorithm2e}
\definecolor{seagreen}{rgb}{0.18, 0.55, 0.24}
\SetAlFnt{\small\color{black}\tt}
\LinesNumbered

\def\BibTeX{{\rm B\kern-.05em{\sc i\kern-.025em b}\kern-.08em T\kern-.1667em\lower.7ex\hbox{E}\kern-.125emX}}

\begin{document}

\title{GradualDiff-Fed: A Federated Learning Specialized Framework for Large Language Model\\}

\author{\IEEEauthorblockN{Amir Faiyaz}
\IEEEauthorblockA{\textit{Department of Computer Science} \\
\textit{Texas Tech University}\\
Texas, USA \\
afaiyaz@ttu.edu}

\and
\IEEEauthorblockN{Tara Salman}
\IEEEauthorblockA{\textit{Department of Computer Science} \\
\textit{Texas Tech University}\\
Texas, USA \\
tsalman@ttu.edu}
}

\maketitle

\begin{abstract}
The rapid proliferation of large language models (LLMs) has created an unprecedented demand for fine-tuning models for specialized domains, such as medical science. While federated learning (FL) offers a decentralized and privacy-preserving approach to collaboratively fine-tune LLMs without sharing raw data, it presents significant challenges, particularly in performance and managing large model sizes efficiently. 
In this paper, we introduce GradualDiff-Fed, an FL framework designed explicitly for LLMs, and their challenge of handling the high parameter size. GradualDiff-Fed reduces communication costs by transmitting only the difference of model weights rather than the entire model during training rounds. Such an approach significantly improves scalability and communication efficiency, making it more feasible to fine-tune LLMs across distributed clients without compromising performance.
Our evaluation demonstrates that GradualDiff-Fed achieves performance on par with centralized training while drastically reducing communication overhead. These results highlight the potential of GradualDiff-Fed as an efficient solution for fine-tuning large models from distributed data in privacy-preserving settings without comprising performance.

\end{abstract}


\section{Introduction}

The recent advancements in large language models (LLMs) \cite{1}, exemplified by systems like ChatGPT and GPT-4, have propelled the field of natural language processing (NLP) to new heights. These models have shown exceptional capabilities, leading to a surge in efforts to fine-tune LLMs for specialized domains such as finance, healthcare, law, and biology \cite{2}. However, fine-tuning LLMs typically demands vast amounts of data, often sourced from diverse, globally distributed entities. This presents a significant challenge when the data in question is sensitive or private \cite{3}. For instance, medical data is highly confidential in the healthcare sector, and sharing it, even with trusted servers, raises serious privacy concerns. Further, sharing the data will raise the risks of non-compliance with stringent data protection laws such as the Health Insurance Portability and Accountability Act (HIPAA) in the U.S. and the General Data Protection Regulation (GDPR) in the European Union \cite{4}. These regulations impose strict guidelines on the handling and sharing personal and sensitive data, making the traditional centralized model of LLM training problematic.

This backdrop underscores the pressing need for distributed learning paradigms that enable LLMs to learn from decentralized data without compromising privacy. Federated learning (FL) emerges as a promising solution, allowing for the collaborative training of models across multiple devices or servers that hold local data samples without requiring the exchange of raw data \cite{5}. By ensuring that data remains on the original devices, FL mitigates the privacy risks associated with centralized model generation and aligns with regulatory requirements such as HIPAA and GDPR \cite{4}.
Despite its potential, research on FL for LLMs remains limited. In this context, FL involves exchanging model updates or gradients rather than raw text data, which helps preserve the privacy of the data while still enabling the language model to benefit from a diverse array of data sources. 

In this paper, we introduce GradualDiff-Fed, an FL framework designed explicitly for fine-tuning LLMs in a distributed fashion while maintaining efficiency and performance. We evaluate GradualDiff-Fed's performance and compare it to traditional and local LLM settings. Our experiments, conducted on medical datasets, demonstrate that GradualDiff-Fed maintains higher accuracy and significantly reduces communication overhead.

The remainder of this paper is organized as follows: Section \ref{background} details the background needed for the paper. Section \ref{System Model} illustrates the overall idea of the proposed GradualDiff-Fed. Section \ref{Exp_analysis} presents the evaluation of GradualDiff-Fed and compares it to traditional techniques. Section \ref{related_works} discusses the related work, and finally, section \ref{conclusion} concludes the paper and discusses future work.

\section{Background}\label{background}
In this section, we provide a brief background on the FL process and traditional LLM paradigm. 
\subsection{Federated Learning (FL) }
FL is a distributed machine learning paradigm that allows local devices to train on their data and generate local model updates to be shared with a central server \cite{5, 6}. These updates can be in the form of model weights or model parameters. The central server then runs a predefined aggregation algorithm to obtain a global model, which will be used for subsequent training by the local devices. In an FL architecture, local devices can be called clients, participants, or workers, and the central server is referred to as the server or aggregator. For simplicity, we use "clients" to represent these local devices and "server" to represent the central server. The FL process is as follows:
\begin{enumerate}
    \item \textbf{Client Selection:}
    A subset of $N$ clients is selected from the total pool of clients to participate in each training round. The selection can be random, based on pre-specified techniques, or per all clients. 
    \item \textbf{Local Model Training:} 
    Each selected client $i$ trains a local model $\theta_i^{(t)}$ on its data by performing gradient descent. For each minibatch $b$, the local model is updated as follows:
    \[
    \theta_i^{(t)} \gets \theta_i^{(t-1)} - \eta \nabla L(b; \theta_g^{(t-1)})
    \]
    where $\eta$ is the learning rate and $L(b; \theta_g^{(t-1)})$ is the loss function for minibatch $b$ given the global model $\theta_g^{(t-1)}$.

    \item \textbf{Local Model Update Sharing:} 
    After training, each client sends its updated model parameters $\theta_i^{(t)}$ to the central server.

    \item \textbf{Global Model Aggregation:} 
    The server aggregates the updates received from all participating clients. A standard aggregation method, called FedAvg\cite{12}, is weighted averaging based on the number of local data samples $n_i$ for each client:
    \[
    \theta_g^{(t)} \gets \frac{1}{N} \sum_{i=1}^{N} \frac{n_i}{\sum_{j=1}^{N} n_j} \theta_i^{(t)}
    \]
    where $\theta_g^{(t)}$ is the updated global model and $n_i$ is the number of samples used by client $i$.

    \item \textbf{Global Model Distribution:} 
    The updated global model $\theta_g^{(t+1)}$ is sent back to the participating clients to begin the next round of local training.
\end{enumerate}

This cycle repeats, with clients continually refining the global model through their local training and updates. It ends when the model coverage or the number of iterations has reached the limit. 

\subsection{Language Modeling}
Language modeling is a core task in NLPs and LLMs and has been employed as a foundational component of state-of-the-art pipelines such as those developed by \cite{7}, \cite{1}. Given a sequence of \( s \) tokens \( \{x_1, x_2, \dots, x_s\} \), the objective of language modeling is to estimate the probability distribution of \( \mathbb{P}(x) \):

  \[ \log \mathbb{P}_\theta (x) = \sum_{i=1}^n \log \mathbb{P}_\theta (x_i \mid x_1, \dots, x_{i-1})
\]

Contemporary neural language models are predominantly powered by architectures such as RNNs \cite{8} or Transformers \cite{9}. These models are characterized by millions or billions of parameters, denoted in this paper as $\theta$. The first step in these models is to transform the input tokens $x$ into a series of vectors via a word embedding matrix $W \in \mathbb{R}^{|V| \times d}$, where $V$ represents the vocabulary and $d$ is the dimension of the hidden layers.
Following the initial transformation, the model computes a hidden representation $h_i$ for each token conditioned on its predecessors:
\[
P_\theta(x_i | x_1, \ldots, x_{i-1}) = \frac{\exp(h_i^\top W_{x_i})}{\sum_{j \in V} \exp(h_i^\top W_j)}.
\]

\section{System Model}\label{System Model}

This section outlines the system model for the LLM utilized in our proposed \textit{GradualDiff-Fed} framework. We use Low-Rank Adaptation (LoRA) as the Parameter-Efficient Fine-Tuning (PEFT) method, which is very effective for large models in resource-limited environments \cite{10,11}.

LoRA, as shown in Algorithm \ref{alg:lora}, breaks the training weights into two parts: a fixed part and a trainable part. The fixed part, called \( \theta \) in this paper, does not change during fine-tuning. A trainable part, called \( \Delta \theta = \Delta \theta_b \Delta \theta_a \), which is a product of two smaller matrices \( \Delta \theta_b \) and \( \Delta \theta_a \).
This setup reduces the number of parameters that need to be updated, making the process faster and more efficient.

\begin{algorithm}
\caption{Low-Rank Adaptation (LoRA)}
\label{alg:lora}
\begin{algorithmic}[1]
\REQUIRE Original weight matrix $\theta$, input data $X$, rank $r$
\ENSURE Adapted weight matrix $\theta' = \theta + A \times B$
\STATE Initialize $A \in \mathbb{R}^{m \times r}$ and $B \in \mathbb{R}^{r \times n}$ randomly
\STATE Freeze the original weight matrix $\theta$
\FOR{each batch of input data $X$}
    \STATE $Y \gets (\theta + A \times B) \times X$
    \STATE Compute loss between $Y$ and ground truth
    \STATE Backpropagate the loss
    \STATE Update $A$ and $B$ while keeping $\theta$ frozen
\ENDFOR
\STATE \textbf{return} $\theta' = \theta + A \times B$
\end{algorithmic}
\end{algorithm}


  
  
      
      
  


We propose \textit{GradualDiff-Fed}, an algorithm inspired by the FedAvg \cite{12} framework. Algorithm~\ref{alg:fed_llm} presents \textit{GradualDiff-Fed}, which is optimized to address the challenges of fine-tuning LLM in FL. While FedAvg aggregates model updates by averaging the full weight parameters from participating clients as was shown in section \ref{background}, \textit{GradualDiff-Fed} introduces key optimizations to enhance scalability and efficiency. First, instead of transmitting the entire updated model (\( \theta_i^{(t)} \)) back to the server, each client in \textit{GradualDiff-Fed} computes and transmits only the difference between its locally updated model and the global model: 
\begin{equation}
\Delta \text{LLM}_i^{(t)} = \theta_i^{(t)} - \text{LLM}_g^{(t)}.
\end{equation}

\begin{algorithm}
\caption{GradualDiff-Fed}\label{alg:fed_llm}
\KwIn{$T$: number of global training rounds, $K$: number of clients, $\text{LLM}$: Large Language Model}
\KwOut{Final global LLM model $\text{LLM}_g^{(T)}$}

\textbf{Initialize:} 
\begin{itemize}
    \item Global LLM model weights $\text{LLM}_g^{(1)}$
    \item Local LLM models $\text{LLM}_i$ for each client $i = 1, 2, \dots, K$
\end{itemize}

\For{each global training round $t = 1, 2, \dots, T$}{
    \tcp{Server sends global LLM model weights to all clients}
    Send global model weights $\text{LLM}_g^{(t)}$ to each client $i$;
    
    \For{each client $i = 1, 2, \dots, K$ \textbf{in parallel}}{
        \tcp{Client fine-tunes or trains LLM on private data}
        Perform local training:
        \[
        \theta_i^{(t)} \gets \theta_i^{(t)} - \eta \nabla L(X_i, Y_i; \theta_i^{(t)})
        \]
        where $\theta_i^{(t)}$ are the LLM model parameters for client $i$ at round $t$, and $\eta$ is the learning rate.
        
        \tcp{Client sends model updates to server}
        Send updated model:
        \[
        \Delta \text{LLM}_i^{(t)} \gets \theta_i^{(t)} - \text{LLM}_g^{(t)}
        \]
        where $\Delta \text{LLM}_i^{(t)}$ is the difference between the locally updated model and the global model.
    }
    
    \tcp{Server aggregates model updates from clients}
    Aggregate updates: 
    \[
    \Delta \text{LLM}_g^{(t)} \gets \frac{1}{K} \sum_{i=1}^{K} \Delta \text{LLM}_i^{(t)}
    \]
    
    \tcp{Server updates global model}
    Update global model:
    \[
    \text{LLM}_g^{(t+1)} \gets \text{LLM}_g^{(t)} + \Delta \text{LLM}_g^{(t)}
    \]
}

\Return Final global LLM model $\text{LLM}_g^{(T)}$
\end{algorithm}

This approach dramatically reduces the amount of data that needs to be sent, making communication much more efficient while keeping all the essential information needed for updating the global model. Second, \textit{GradualDiff-Fed} improves how updates are combined on the server by averaging only the differences (\( \Delta \text{LLM}_i^{(t)} \)) sent by the clients. This reduces the processing and storage requirements on the server. The global model is updated step by step using the following formula:
\begin{equation}
\text{LLM}_g^{(t+1)} = \text{LLM}_g^{(t)} + \frac{1}{K} \sum_{i=1}^{K} \Delta \text{LLM}_i^{(t)}.
\end{equation}

This step-by-step update process makes handling large models with billions of parameters easier and more efficient. By solving the challenges of FedAvg for handling large models, \textit{GradualDiff-Fed} provides a scalable and efficient solution for FL without losing model quality.

\section{Performance Evaluation}\label{Exp_analysis}
In this section, we discuss our experimental setup 
including data and model settings. We also discuss the evaluation results and comparisons. 

\subsection{Experimental Setup}

\textbf{FLSetup:} We implemented a custom FL system without using pre-existing frameworks. The training was performed on an NVIDIA GeForce RTX 3090 GPU with a compute capability of 8.6. The FL system consisted of a centralized server and five clients. The base model was \textit{llama2\_7b}, a 7 billion parameter size with 4k content size, fine-tuned with LoRA alg~\ref{alg:lora}. We applied LoRA to all global layers within the adapters. For optimization, we employed AdamW. We searched for the optimal learning rate within the following range: \(\{1 \times 10^{-5}, 3 \times 10^{-5}, 5 \times 10^{-5}, 8 \times 10^{-5}, 1 \times 10^{-4}\}\).  We applied a 4-bit quantization technique. This reduced the model size significantly while maintaining a balance between performance and accuracy. The \textit{mental\_health\_chatbot\_dataset} with a token size of 100,000, was used as a dataset to provide domain-specific training for mental health chatbot applications. We conducted experiments for 15 rounds across three configurations: FL, centralized learning, and local-client setting. The summary of parameters is presented in Table~\ref{tab:training_params}.

\begin{table}
\centering
\caption{Training Parameters for Model}
\label{tab:training_params}
\begin{tabular}{|l|c|}
\hline Gradient Clipping (Max Gradient Norm) & 0.3          \\ \hline
Weight Decay                       & 0.001          \\ \hline
Optimizer                          &  AdamW  \\ \hline

Warmup Ratio                       & 0.03           \\ \hline
LoRA Attention Dimension ($r$)     & 64             \\ \hline
LoRA Alpha                         & 16             \\ \hline
LoRA Dropout                       & 0.1            \\ \hline
\end{tabular}
\end{table}

\subsection{Performance Metrics}
We evaluated our FL system using four key metrics: training loss, perplexity, BLEU score, and computation time.
 \textbf{Training Loss} measures how well the model minimizes the difference between predicted and actual values. It is calculated for FL, local clients, and centralized learning to evaluate model convergence.
\textbf{Perplexity} assesses how well a model predicts a sequence, defined as the exponentiation of the average negative log-likelihood of a sequence:
\begin{equation}
    \text{Perplexity}(X) = \exp\left( -\frac{1}{T} \sum_{i=1}^{T} \log p_\theta(x_i \mid x_{<i}) \right)
\end{equation}
    where \( p_\theta(x_i \mid x_{<i}) \) represents the probability of the \(i\)-th token given preceding tokens. Lower perplexity indicates better model performance.
\textbf{BLEU Score} evaluates the alignment of machine-generated text with reference text by calculating n-gram overlap precision while penalizing brevity. Higher BLEU scores indicate better text quality. Finally, \textbf{Computation Time} tracks the duration of one training round for FL, local client, and centralized learning, providing insights into computational efficiency.

\subsection{Performance Evaluation}
\begin{figure}
  \centering
  \includegraphics[width=0.90\linewidth]{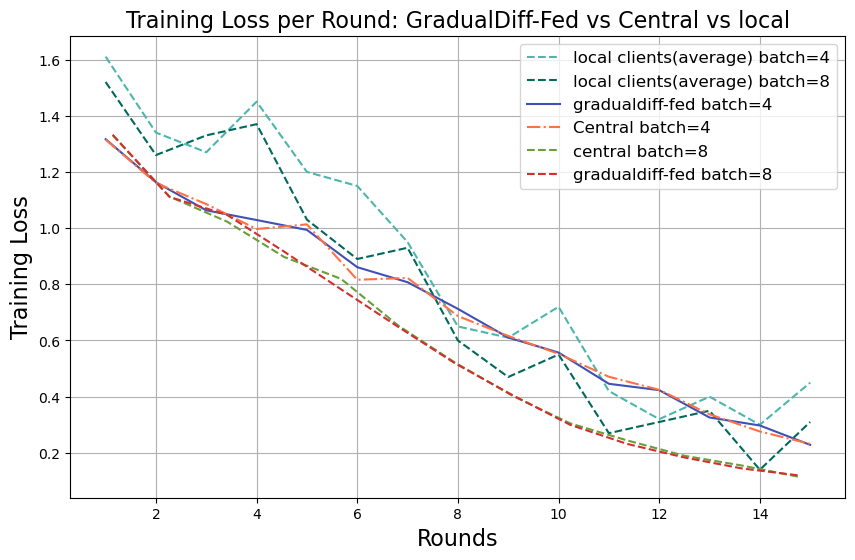}
  
  \caption{Training loss across 15 rounds comparing local client(average), GradualDiff-Fed and Central approaches for different batch sizes (Batch=4 and Batch=8)}
  \label{fig:FL}
\end{figure}

Our results show that GradualDiff-Fed performs almost as accurately as centralized learning. As shown in Figure~\ref{fig:FL}, GradualDiff-Fed achieves comparable training loss to the centralized approach across different batch sizes. From Table~\ref{tab:comparison_bit_precision}, we can see that for batch size 4, GradualDiff-Fed achieves a final loss of 0.22, outperforming centralized training (0.31) by 29.0\%, while local training degrades to 0.45 due to severe overfitting. With batch size 8, GradualDiff-Fed converges to 0.11 loss, surpassing centralized training (0.15) by 26.7\%, whereas local training stagnates at 0.31, reflecting its inability to generalize beyond narrow data subsets. However, Local client training experiences persistent loss convergence challenges, primarily due to data distribution limitations. Each client operates on a subset of the overall dataset, resulting in fewer data points per local model. Without proper aggregation or averaging techniques, individual clients are prone to overfitting, leading to poor performance across the broader data distribution.

GradualDiff-Fed’s effectiveness is driven by two key mechanisms. First, delta parameterization: $\Delta LLM_i^{(t)} = \theta_i^{(t)} - LLM_g^{(t)}$ keeps updates consistent with the global model, reducing interference while handling data heterogeneity. This helps achieve 15–30\% lower losses compared to centralized training. Second, synchronous aggregation with all clients participating ensures steady updates by avoiding participation gaps, leading to smooth loss reduction. For example, with a batch size of 8, the FL loss drops to 0.11, beating centralized training’s 0.15.

\begin{table}
\caption{Training Loss Across Bit Precision (BP), Batch Size (BS), and Scenarios}
\label{tab:comparison_bit_precision}
\begin{tabular}{|c|c|ccc|ccc|}
\hline
\multicolumn{2}{|c|}{\textbf{Config}} & \multicolumn{3}{c|}{\textbf{Initial Loss}} & \multicolumn{3}{c|}{\textbf{Final Loss (Round 15)}} \\
\cline{1-8}
\textbf{BP} & \textbf{BS} & \textbf{FL} & \textbf{Central} & \textbf{Local} & \textbf{FL} & \textbf{Central} & \textbf{Local} \\
\hline
4-bit & 4 & 1.36 & 1.31 & 1.61 & 0.22 & 0.31 & 0.45 \\
\hline
4-bit & 8 & 1.33 & 1.39 & 1.52 & 0.11 & 0.15 & 0.31 \\
\hline
\end{tabular}
\end{table}

In Table~\ref{tab:comparison_scores}, the Central and GradualDiff-Fed models exhibit identical BLEU scores (0.55), indicating equivalent surface-level n-gram overlap with reference outputs. The Central model's lower perplexity (15.36 ± 0.2 vs. 16.78) through optimized cross-entropy minimization across the full data distribution, whereas GradualDiff-Fed's marginally higher perplexity (16.78) reflects the inherent challenge of modeling non-IID client data partitions during federated optimization. The Local (average) model's degraded metrics - 19.2 perplexity (+25.3\% vs. Central) and BLEU = 0.47 (-14.5\% vs. Central) - expose the limitations of isolated training on narrow token distributions.

\begin{table}[]
\centering
\caption{Comparison of Perplexity and BLEU Score Across Central, GradualDiff-Fed and Local(average) Approaches}
\label{tab:comparison_scores}
\begin{tabular}{|l|c|c|c|c|c|}
\hline
\textbf{Metric} & \textbf{Central} & \textbf{FL} &\textbf{Local}\\
\hline
Perplexity      & 15.36            & 16.78    & 19.20   \\
BLEU Score      & 0.55             & 0.55      & 0.47 \\
\hline
\end{tabular}
\end{table}

The computational timing results in Table~\ref{tab:timing_comparison} show that GradualDiff-Fed is significantly faster than Centralized Learning by reducing communication costs through weight difference updates, achieving a 36\% reduction in per-round time (4.3 vs. 6.7). While Local-Client Learning is the fastest (4.1) due to the absence of synchronization overhead, it lacks collaborative aggregation, resulting in severe overfitting (e.g., 0.45 final loss for BS=4 vs. GradualDiff-Fed’s 0.22). On the other hand, Centralized Learning though robust, suffers from high computational overhead due to full-dataset processing. Our framework enhances efficiency and accuracy by using delta parameterization to send only weight differences, reducing communication overhead while preserving centralized performance.
\begin{table}
\centering
\caption{Computational Timing for Different Learning Approaches}
\label{tab:timing_comparison}
\begin{tabular}{|l|c|}
\hline
\textbf{Approach} & \textbf{Time per Training Round } \\
\hline
Federated Learning & 4.3 \\
Centralized Learning & 6.7 \\
Local-Client Learning & 4.1 \\
\hline
\end{tabular}
\end{table}

\section{Related Works}\label{related_works}
Fine-tuning federated-LLM has gained significant attention. Previous work,\cite{ro-etal-2022-scaling} focused on mitigating computation bottleneck by utilizing k-level uniform quantization and partial model training.
FedKD \cite{23} improved communication efficiency through adaptive mutual knowledge distillation but faces the challenge of increased client-side computational overhead due to the knowledge exchange between smaller and larger models. FwdLLM \cite{xu2024fwdllmefficientfedllmusing} utilized backpropagation free-training of LLM by relying solely on perturbed inference. Their approach leveraged parameter-efficient training \cite{10,11} and adaptive load allocation. FedPEFT \cite{18} reduced communication by locally tuning model weights, while FedIT \cite{zhang2024buildingfederatedgptfederated} faced noise from naive LoRA averaging, and FedFLORA \cite{wang2024flora} mitigated this with a stacking-based aggregation. The authors in \cite{zerothorder} integrated the memory-efficient zeroth order optimization within FL for LLM, accelerated convergence, and improved loss reduction with a personalized learning rate. 

This work differentiates from previous studies by introducing GradualDiff-Fed, which reduces communication overhead by having clients transmit only the weighted differences between their local updates and the global model.

\section{Conclusion}\label{conclusion}
FL plays a pivotal role in privacy-preserving distributed machine learning in medical contexts. However, its use for LLMs is still limited. 
We introduced GradualDiff-Fed, a scalable framework for fine-tuning LLMs in federated settings. By transmitting only model weight differences rather than the full model, GradualDiff-Fed significantly reduces communication overhead while maintaining performance comparable to centralized baselines. Results showed on par with centralized learning, with similar BLEU scores and perplexity indicating comparable text quality. The framework’s communication efficiency provides a strong foundation for future enhancements. Future work should check GradualDiff-Fed performance under non-IID setting. Building on these improvements through advanced optimization strategies such as parameter-efficient transfer learning and domain adaptation can refine performance and applicability in real-world application scenarios. Further, privacy leakage with GradualDiff-Fed by analyzing the model differences and its defenses is yet to be explored.


\bibliographystyle{IEEEtran}
\bibliography{ref.bib}

\vspace{12pt}

\end{document}